\documentclass[sigconf]{acmart}
\usepackage{multirow}
\usepackage{booktabs}
\usepackage{graphicx}
\usepackage{subfig}
\usepackage{tablefootnote}
\usepackage{array}
\usepackage[ruled,vlined]{algorithm2e}
\usepackage{bbm}
\usepackage{amsmath}
\PassOptionsToPackage{normalem}{ulem}
\usepackage{ulem}
\usepackage{mathrsfs} 
\usepackage{mathtools}
\usepackage[utf8]{inputenc}
\DeclareMathOperator*{\argmax}{arg\,max}

\AtBeginDocument{%
  \providecommand\BibTeX{{%
    \normalfont B\kern-0.5em{\scshape i\kern-0.25em b}\kern-0.8em\TeX}}}

\setcopyright{acmcopyright}
\copyrightyear{2022}
\acmYear{2022}

\acmConference[DRL4IR@SIGIR]{The 3rd Workshop on Deep Reinforcement Learning for Information Retrieval at SIGIR'22}{July 15, 2022}{Madrid, Spain}

\begin{document}

\title{BCRLSP: An Offline Reinforcement Learning Framework\\ for Sequential Targeted Promotion}

\author{Fanglin Chen}
\email{flchen@bus.miami.edu}
\affiliation{%
  \institution{University of Miami}
  \city{Miami}
  \state{FL}
  \country{USA}
}

\author{Xiao Liu}
\email{xliu@stern.nyu.edu}
\affiliation{%
  \institution{New York University}
  \city{New York}
  \state{NY}
  \country{USA}
}

\author{Bo Tang}
\email{tangbo.t@alibaba-inc.com}
\affiliation{%
  \institution{Alibaba Group}
  \city{Hangzhou}
  \country{China}
}

\author{Feiyu Xiong}
\email{feiyu.xfy@alibaba-inc.com}
\affiliation{%
  \institution{Alibaba Group}
  \city{Hangzhou}
  \country{China}
}

\author{Serim Hwang}
\email{serimh@andrew.cmu.edu}
\affiliation{
    \institution{Carnegie Mellon University}
    \city{Pittsburgh}
    \state{PA}
    \country{USA}
}

\author{Guomian Zhuang}
\email{guomian.zgm@alibaba-inc.com}
\affiliation{%
  \institution{Alibaba Group}
  \city{Hangzhou}
  \country{China}
}

\renewcommand{\shortauthors}{Chen, et al.}


\begin{abstract}
  We utilize an offline reinforcement learning (RL) model for sequential targeted promotion in the presence of budget constraints in a real-world business environment. In our application, the mobile app aims to boost customer retention by sending cash bonuses to customers and control the costs of such cash bonuses during each time period. To achieve the multi-task goal, we propose the Budget Constrained Reinforcement Learning for Sequential Promotion (BCRLSP) framework to determine the value of cash bonuses to be sent to users. We first find out the target policy and the associated Q-values that maximizes the user retention rate using an RL model. A linear programming (LP) model is then added to satisfy the constraints of promotion costs. We solve the LP problem by maximizing the Q-values of actions learned from the RL model given the budget constraints. During deployment, we combine the offline RL model with the LP model to generate a robust policy under the budget constraints. Using both online and offline experiments, we demonstrate the efficacy of our approach by showing that BCRLSP achieves a higher long-term customer retention rate and a lower cost than various baselines. Taking advantage of the near real-time cost control method, the proposed framework can easily adapt to data with a noisy behavioral policy and/or meet flexible budget constraints.
\end{abstract}



\maketitle

\section{Introduction}
\label{sec:intro}
In order for a mobile app to be successful, it must have loyal users that remain active on the app over time. One important digital marketing tool to boost user retention is sequential targeted promotion. On one hand, \emph{targeted} promotion has been widely applied by mobile apps as they have more access to customers' information and have a better understanding of their behavior in the era of big data. On the other, \emph{sequential} promotion
can help mobile apps build customer loyalty and increase customer lifetime value over time, ultimately boosting profits in the long run. One form of sequential targeted promotion, which combines the two kinds of promotion above, is to send personalized cash bonuses sequentially to users if they keep using the app. 

In this paper, we design a policy to allocate such cash bonuses throughout the lifecycle of each customer, aiming to maximize user retention while satisfying the budget constraint during each time period. 
We use the reinforcement learning (RL) framework to formulate the problem of sequential targeted promotion as a Markov decision process (MDP). Reinforcement learning algorithms enable us to dynamically adjust the cash bonuses over time for each customer, in order to maximize long-run user engagement. Sequential targeting is superior to static targeting because a customer's reaction to a given cash bonus may depend not only on the current cash bonus but also on all the previously-received cash bonuses. Therefore, the app's decision on the value of cash bonuses can generate long term consequences instead of only immediate outcomes.

In the e-commerce setting, the target policy should be trained and tested in an offline manner from logged data for two major reasons. First, online training relies on real-time interactions with customers, and it is often prohibitively expensive to collect enough data needed to train effective models. Offline data, in contrast, is natural to record as the interactions are already taking place in the existing environment. Second, during online training, the proposed policy interacts with real users without being evaluated in advance. Therefore, the firm has little control over the online process, such that the model performance could be even worse than the behavioral policy, resulting in lower profits and/or higher costs. 

There are two major technical challenges in applying offline RL to sequential targeted promotion. One challenge results from the distributional shift of actions due to the discrepancy between the behavior policy and the target policy. This is a fundamental challenge in almost all deployments of offline RL~\cite{levine2020offline}. The other challenge is that the problem is multi-task, i.e., both the optimality and feasibility (i.e., that costs cannot exceed the budget) need to be assessed before the policy can be deployed to real users.

In this paper, we focus on sequential targeted promotion with constraints in each time period, in order to fulfill flexible budget constraints that mimic the real-world industrial settings. To address the distributional shift issue, we use a batch constrained model-free RL model, specifically, batch constrained learning~\citep{fujimoto2019off, fujimoto2019benchmarking}.
We choose model-free solutions over model-based solutions because customer behavior in an information-rich environment is too complicated to model or simulate without any bias. To address the multi-task challenge, we incorporate budget constraints by adding a linear programming model to maximize the Q-values learned from the RL model given the budget constraints. 

The main contributions of our work are summarized as follows:
    
    
     

\begin{itemize}
    
    \item We formalize the problem of allocating sequential targeted cash bonuses among customers and propose a two-stage framework, Budget Constrained Reinforcement Learning for Sequential Promotion (BCRLSP), to solve it. Specifically, we combine an offline RL model with a near real-time LP model to generate a robust policy under the budget constraints.
     
    \item The BCRLSP framework we proposed is verified empirically through both offline and online experiments. The results of offline experiments show that our solution consistently outperforms the baseline models in boosting customer retention and controls promotion costs. Online A/B tests also demonstrate the effectiveness of our method in real-life applications.
    
    \item Since the RL model requires significantly more running time and computational resources than the LP model, we train the RL model using logged offline data but solve the LP model with online data. In this way, the target policy needs to be trained only once but can be adjusted to satisfy any budget constraints flexibly.
\end{itemize}

\section{Related Literature}
\label{sec:related literature}
\subsection{Reinforcement Learning in Digital Marketing}
\label{sec:reinforcement learning for sequential targeted promotion}
In recent years, we have witnessed the rapid development of RL techniques and widespread applications of RL. In digital marketing, RL is expected to revitalize the industry and modernize various operations. For example, prior research has applied RL to solve digital marketing problems related to search \cite{wei2017reinforcement, xia2017adapting, xiao2019aliisa, hu2018reinforcement}, recommendation \cite{zhao2018deep, cai2018reinforcement, theocharous2015personalized, zhao2018recommendations}, online advertising \cite{yang2016dynamic, zhao2018deep1, jin2018real, cai2017real, wu2018budget, schwartz2017customer}, and pricing~\citep{misra2019dynamic}.
 
However, most prior works in this stream of literature only focus on a single objective. For example, budget is usually not involved in the applications of search or recommendation. Our paper differentiates from them by solving a multi-task problem: return maximization under constraints. Although some RL applications in online advertising also consider the budget constraint, for instance, \citet{cai2017real} and \citet{wu2018budget} use RL to solve the budget-constrained real-time-bidding problem,they do not allow for offline training to make sure the costs of the proposed policy are controlled. 

The most relevant work to our study is \citet{xiao2019model}, which uses a model-based MDP with a global constraint in sequential targeted promotion. Our work differentiates from \citet{xiao2019model}  in two major aspects. First, the state space is in high dimension, estimating the transition dynamics is very challenging, so we choose model-free algorithms over the model-based ones. Second, the CMDP setting controls the average budget over the whole trajectory of each user, while budget constraint in BCRLSP is set as the average cost per time step, which is more flexible and easier for the company to adjust.


\subsection{Reinforcement Learning Algorithm with Constraints}
\label{sec:reinforcement learning algorithm with constraints}
In standard reinforcement learning (RL), a learning agent seeks to optimize the long-term return, which is based on a real-valued reward signal (\citet{sutton1998introduction}). However, in our setting, we want to ensure reasonable performance and at the same time respect budget constraints during deployment. RL algorithms with constraints are mostly formulated as a constrained Markov Decision Process (CMDP) introduced by~\citet{altman1999constrained}.
In CMDPs, the agent's goal is to maximize the long-run rewards while satisfying some linear constraints on the long-run costs. \citet{altman1999constrained} give an LP-based solution under the assumption that the CMDP process has known dynamics and finite states.

Based on that, some model-free approaches are also proposed to solve RL with constraints in high-dimensional settings. \citet{achiam2017constrained} use constrained policy optimization (CPO), focus on online safe exploration, and solve CMDP for continuous actions during the learning process. \citet{le2019batch} describe a batch offline algorithm with PAC-style guarantees for CMDPs. Later,~\citet{miryoosefi2019reinforcement} generalize this algorithm that can work with arbitrary convex constraints.


Although CMDP is the mainstream framework in RL with constraints, the long-run cost constraint in CMDP cannot satisfy the requirements of most business applications. CMDP models consider cost as part of the return and maximize the accumulated return. Mostly, companies have a budget constraint for each period of time, a week or a month. It's difficult to define the value of the constraints for the long-run costs in this circumstance.
In contrast, our BCRLSP method can meet the dynamic budget requirement set by the company by applying constrained optimization after solving the RL model. Our constraint is set per time period, which is more flexible and easier for the company to adjust. In this case, we are able to have exact control over the budget in near real-time.

\section{Context}
\label{sec:context}
\begin{figure}[t]
\centering
\includegraphics[width=0.5\textwidth]{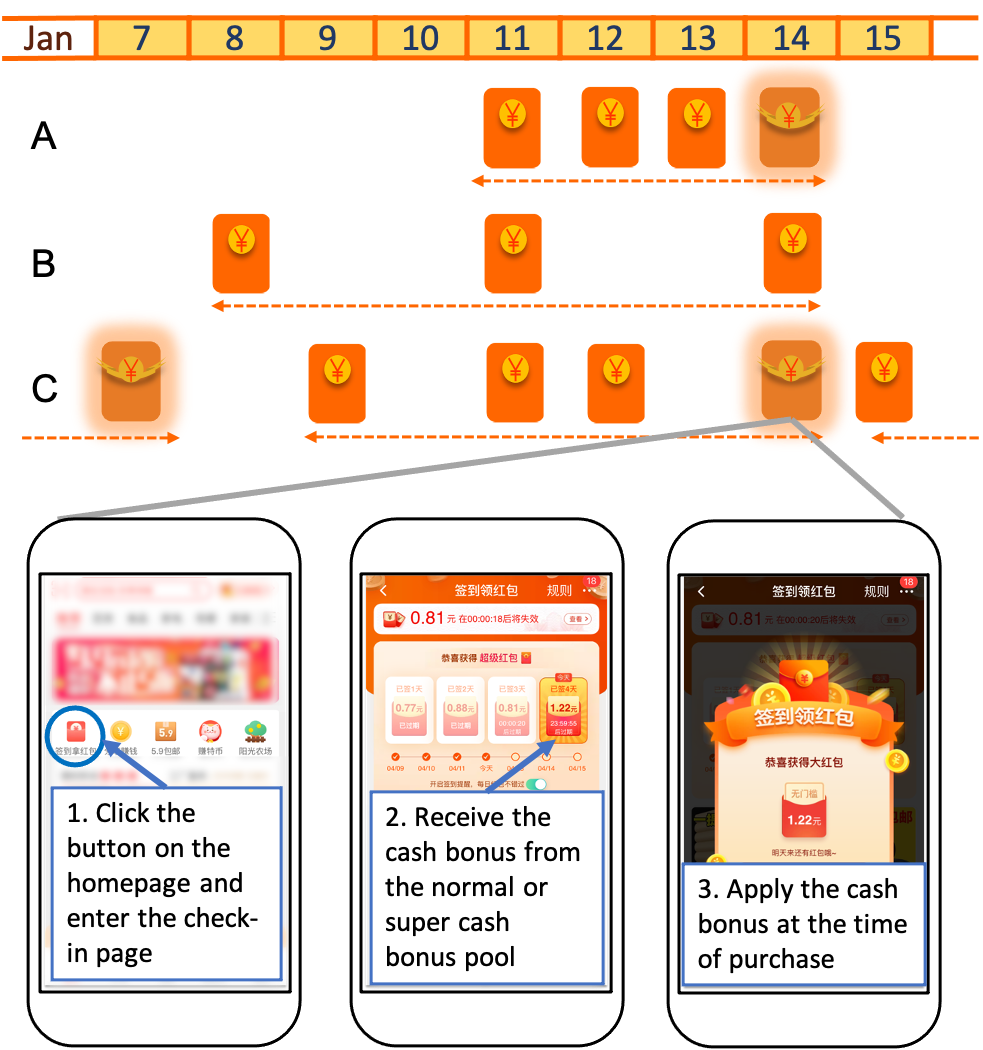} 
\caption{Cycles and cash bonuses in Taobao Special Offer Edition}
\label{fig:problem visualisation}
\end{figure}

The context of our problem is the daily check-in cash bonus strategy used by Taobao Special Offer Edition. It is a mobile shopping app launched by Alibaba in 2019 and focuses on ultra-competitive product pricing. 
The mechanism of this strategy is shown in the lower part of Figure~\ref{fig:problem visualisation}.
When a user opens the Taobao Special Offer Edition app, he/she can click a button and enter the daily check-in page. A cash bonus can be claimed after clicking the check-in button. The corresponding cash bonus can be subtracted from the final payment at the time of purchase within 24 hours. For each user, only one cash bonus can be claimed on the same day. The app needs to design a policy to optimally allocate the cash bonuses to each consumer on each day they log in. 

Two important features of the setting make the problem a constrained optimization problem. First, the average cash bonus for each consumer distributed on each day is limited. 
This daily quota is clearly predefined by operational business units for economic reasons. Models without such a constraint may bring substantial economic losses as a large amount of abnormally high-value cash bonuses may be sent out. Also, such a constraint makes the A/B test meaningful and comparable. Undoubtedly, if two policies are under different budget constraints, the one with a larger budget could more easily create higher user retention. Therefore, a fair comparison of different policies should impose the same constraint.

Second, the set of available cash bonuses is discrete and constrained. 
One cycle is at most seven days long, and the cycle restarts once the user collects four cash bonuses in the cycle. In each cycle, the available cash bonuses vary by day. In order to encourage users to log in to the app frequently, the fourth cash bonus in a cycle will be larger than the first three. 
There are two sets of cash bonuses from which the value of cash bonuses is chosen. One is for normal cash bonuses and the other is for super (i.e., large) cash bonuses. Users that are on the first three days of the cycle may only receive a normal cash bonus, whereas users that are on the fourth day of the cycle may only receive a super cash bonus. Note that this rule can be regarded as public information and we may assume that users know the rule well.
We show some examples of possible cycles in sequences A, B, and C in the upper part of Figure~\ref{fig:problem visualisation}, with the super cash bonuses highlighted.


\section{Model}
\label{sec:model}
In this section, we describe the details of the BCRLSP framework and the baseline models.
To generate a robust policy under the budget constraints, our proposed BCRLSP consists of two modules, an offline RL model followed by a near real-time LP model.

Due to this combination of two modules, BCRLSP can not only solve the constrained optimization problem efficiently but also adapt to the flexible budget constraints in business. Mostly, companies have a budget constraint for each period, a week or a month. Our BCRLSP method can meet the dynamic budget requirement set by the company by applying constrained optimization after solving the RL model. Our constraint is set per time period, which is more flexible and easier for the company to adjust. Specifically, we are able to have exact control over the budget in near real-time.

\begin{figure*}[t]
\centering
\includegraphics[width=0.95\textwidth]{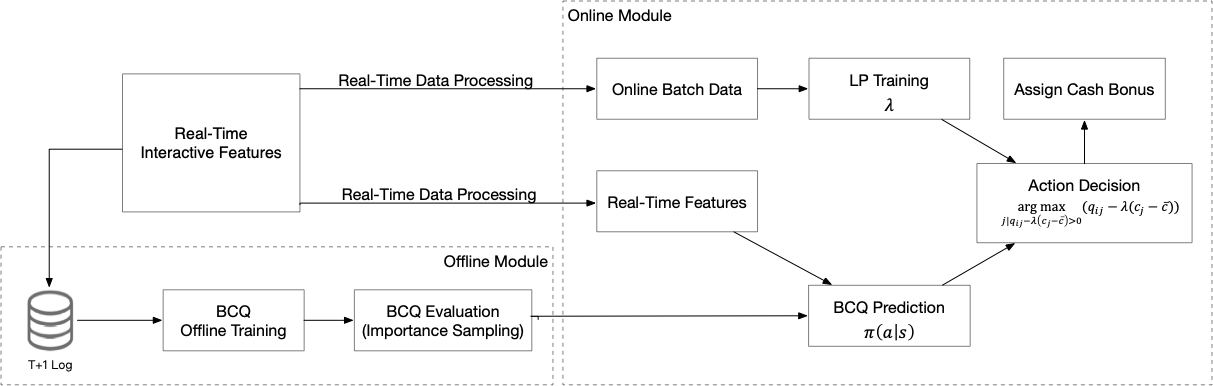} 
\caption{BCRLSP Model framework}
\label{fig:model framework}
\end{figure*}

The model framework is shown in Figure~\ref{fig:model framework}. 
As it is a daily sign-in problem, the data are updated daily. Specifically, we first train the RL module with the data in an offline manner. For all the users who have login records in the platform history, we run the RL model to predict the next (action, Q-value) pair for the user, where the action means the amount of cash bonus to be assigned to the user if he/she logs the next day. Then in the online setting, each time a user interacts with the platform, we have a new record in the online data, which is the (action, Q-value) pair of the user predicted by offline RL. With the online data, the linear programming module is trained to update the policy under the constraints of the required per person budget. 

\subsection{MDP formulation}
\label{sec:mdp}
In this research, we aim to learn a policy from offline data such that when deployed online, it maximizes the cumulative user retention rate while still keeping costs under constraints on the Taobao Special Offer Edition. We address this problem with a forward-looking sequential targeting method based on the model-free deep reinforcement learning algorithm. A learning agent chooses a value of cash bonus under a policy, and the value will be allocated to the user once claimed. We define the act of choosing the value of cash bonuses as an \emph{action}. We formulate the sequential targeted promotion problem as a Markov Decision Process (MDP).

Interactions between the agent and the environment (users collecting the cash bonuses) are recorded in logged offline data $\mathcal{D}$, which is a set of trajectories $\tau_i$ as follows: Let $T$ be the maximum length of time step as far as we concern.\footnote{ We consider a cycle of user behavior as a trajectory and restart a trajectory when a cycle ends, so $T=4$ in our context.}  
A trajectory is a sequence of user states, agent actions, and rewards, 
$\tau_i=\{(s_t^i,a_t^i,r_t^i)\}_{t=1}^{T}$ for time step t and customer $i$.\footnote{For simplicity, in the rest of the paper, we denote a trajectory as $\tau$ without the subscript $i$.} State $s_t^i$ is the user characteristics, action $a_t^i$ is the value of cash bonuses, and reward $r_t^i$ is a 0-1 value indicating whether or not the user logs in the next day. Note that $\mathcal{D}$ is collected from the behavior policy $\pi_b$ that interacted with the real users in the past and that action $a_t^i \sim \pi_b(\cdot \mid s_{t}^i)$. For each action $a_t^i$, it has a corresponding cost $c_t^i$.

\subsection{Batch Constrained Deep Reinforcement Learning}
\label{sec:batch constrained deep reinforcement learning}
The first part of our model is reinforcement learning, which is solved by the Batch-Constrained Q-learning (BCQ) algorithm in the discrete action space~\cite{fujimoto2019benchmarking}.
We choose to use BCQ in an effort to restrict the action space and force the agent towards behaving close to the behavior policy. We do this for economic concerns as this is a cost-sensitive problem. We train the default BCQ model from~\cite{fujimoto2019benchmarking} using the offline data to estimate the expected return of all state-action pairs.

\paragraph{User Behavior Model} Given users' trajectories
$\mathcal{D} =  \{\tau_1\,\dots,\tau_{n}\}$,
we use a neural network $\mathcal{U}_{\omega}$ with a parameter $\omega$ to model the unobserved behavioral policy that generates the offline data. 

\paragraph{Agent} At each time step, the reinforcement learning agent faces state $s$, takes action $a$, receives reward $r$, and transits to next state $s'$. The agent aims to maximize the discounted sum reward (i.e., return) $R_t=\sum_{i=t+1}^{T}\gamma^i r(s_i, a_i, s_{i+1})$, where $\gamma \in [0,1) $ is the discount factor. The agent takes the logged feedback data $\mathcal{D}$ as input and outputs the optimal policy $\pi^*(a|s)$ and the corresponding optimal value function $Q^*(s,a)$ that maximize the return $R_t$. In each iteration of the training process, the action is selected with the highest $Q$ value among the candidate actions whose relative probability is above a certain threshold in the behavioral policy. Specifically, the chosen action is characterized as below:

\begin{equation}
\label{eq:select action}
    \pi(a\mid s)= {\argmax}{arg\,max}_{a\mid\mathcal{U}_{\omega}(a\mid s)/\max_{\hat{a}}\mathcal{U}_{\omega}(\hat{a}\mid s)>\xi}Q_{\theta}(s,a)
\end{equation}
where $\xi$ is the threshold probability that an action has to satisfy in order to be considered, and $Q_{\theta}$ is the value function approximated by a neural network with a parameter $\theta$. Note that $a$ is selected only among those that satisfy $\mathcal{U}_{\omega}(a\mid s)/\max_{\hat{a}}\mathcal{U}_{\omega}(\hat{a}\mid s)>\xi$, i.e., the probability of $a$ needs to exceed a threshold $\xi$ relative to the most probable action $\hat{a}$ in the behavioral policy. If we set the threshold $\xi = 0$, then the model would be the normal deep Q-learning model. For $\xi = 1$ the model will be a simulator of the actions contained in the batch.
Next, we update the value function $Q_{\theta}$ through Q-learning~\citep{watkins1989learning}. Specifically, we minimize the loss function on mini-batches as below:

\begin{equation}
\begin{split}
\label{eq:Q-learning}
    & \mathcal{L}(\theta) = l_{\kappa}\biggl( r        \nonumber \\
    & + \gamma \max_{a'|\pi_b(a'|s')/\max_{\hat{a}}\pi_b(\hat{a}|s')>\xi}Q_{\theta'}(s',a')\nonumber \\
    & -Q_{\theta}(s,a) \biggr)
\end{split}
\end{equation}
where $l_{\kappa}$ denotes the Huber loss~\citep{huber1964robust}:

\begin{equation}
\label{eq:Huber loss}
l_{\kappa}(\delta)=
\begin{cases}
0.5\delta^2 & \text{if } \delta \le \kappa\\
\kappa(|\delta|-0.5\kappa) & \text{otherwise }
\end{cases}.
\end{equation}

\subsection{Near Real-Time Linear Programming}
\label{sec:online linear programming}
The second part of our model is a linear programming model followed by the RL module. 
It is natural to use the LP method for incentive allocation problems in a business setting to solve the large-scale constrained optimization problem efficiently. Under the given budget constraints, to maximize the return, i.e., the long-term retention of users, we use Q-values of actions learned from BCQ as the input of the LP model.

Q-values are a result of the unconstrained RL algorithm. Since Q-values represent the expected return $R_t$ starting from state s, following policy $\pi$, and taking action $a$, we are essentially maximizing the total return across all customers while satisfying the budget constraint in the LP model. 

For a group of $N$ customers, each customer, $i$, is assigned to one and only one level of a cash bonus $j$ (i.e., action) among $M$ levels. The objective is to maximize the Q values of all customers under the constraint of the average cost. The problem formulation is as below:

\begin{align}
\label{eq:linear programming}
    {\argmax}{arg\,max}_{x_{ij}} \ & \sum_{i=1}^{N}\sum_{j=1}^{M}q_{ij}x_{ij} \nonumber\\
    \text{s.t.}\quad 
    \sum_{i=1}^{N}\sum_{j=1}^{M}c_{j}x_{ij} &\leq \sum_{i=1}^{N}\sum_{j=1}^{M}\bar{c}x_{ij} \nonumber\\
    \sum_{j=1}^{M}x_{ij} &= 1, \quad\forall i=1,...,N \nonumber\\
    x_{ij} &\in \{0,1\}
\end{align}
where $x_{ij}$ denotes whether customer $i$ receives action $j$, $q_{ij}$ is the Q value corresponding to the state of customer $i$ and action $j$, $c_j$ is the cost incurred from action $j$, and $\bar{c}$ is the average cost externally set by the company. 

It is worth noting that our budget constraint is the average cost per customer per day, so the total budget is not fixed but rather depends on the number of users who log in on a specific day. This setting fits our goal to increase the number of active users. Suppose we set an upper limit to the total budget, then as the number of users increases, the average cash bonus each user gets will drop, and it will be harder and harder to keep new customers active.

By introducing a Lagrange multiplier $\lambda$, we have the dual problem:

\begin{align}
\label{eq:dual problem}
    \min_{\lambda} \max_{x_{ij}} \  \sum_{i=1}^{N}\sum_{j=1}^{M}q_{ij}x_{ij} &- \lambda\left(\sum_{i=1}^{N}\sum_{j=1}^{M}c_{j}x_{ij}-N\bar{c}\right) \nonumber\\
    \text{s.t.} \quad 
    \sum_{j=1}^{M}x_{ij} &= 1, \quad \forall i=1,...,N \nonumber\\
    x_{ij} &\in \{0,1\} \nonumber\\
    \lambda &\geq 0
\end{align}

Note that the formulation above approximates the 0-1 integer program in Equation~\ref{eq:linear programming}, which is acceptable in exchange for the efficiency of solving this problem. We further transform the dual problem above into a problem only with respect to $\lambda$:

\begin{align}
\label{eq:transformed dual problem}
    \min_{\lambda} \ \sum_{i=1}^{N}\max_{1 \leq j \leq M}\{q_{ij} & -\lambda c_{j}\} + \lambda (N\bar{c}) \nonumber\\
    \text{s.t.} \quad 
    \lambda &\geq 0
\end{align}

After solving for $\lambda$, the $x_{ij}$ in the original problem is given as:

\begin{equation}
\label{eq:linear programming solution}
    x_{ij}=1 \quad \text{when} \quad j= {\argmax}{arg\,max}_{j|q_{ij}-\lambda(c_{j}-\bar{c})\geq 0} (q_{ij}-\lambda(c_{j}-\bar{c}))
\end{equation}

The $\lambda$ parameter reflects how sensitive users are to cash bonuses. If $\lambda$ is large, it means that users' retention rates do not change much as cash bonuses increase. Thus, the optimal strategy is to assign small cash bonuses to users, which will not affect the retention rates a lot but significantly decrease the costs. On the contrary, a small $\lambda$ indicates that uses are very sensitive to cash bonuses, and we need to send large cash bonuses to make sure they continue to use the app.

The key to solving the LP problem is to calculate $\lambda$ such that (1) it accurately reflects the distribution of user sensitivity and (2) it is updated in a short time. To achieve the first goal, it is necessary that the distribution of user sensitivity is stable over time. Since we employ the $\lambda$ values solved using historical data, we use the logged data in the past 24 hours up to the current time point to calculate $\lambda$, rather than the logged data in a shorter time period. The distribution of user sensitivity on the previous day will be similar to that on the current day, enabling us to acquire more accurate $\lambda$ values and have better control of costs.

As for the second goal, we update the logged data and recalculate the corresponding $\lambda$ every 10 minutes. Theoretically, we can update $\lambda$ more frequently or even in real-time, and using a shorter time interval also gives more accurate results. But increasing the frequency increases computational time, and we find that a time interval of 10 minutes achieves the best balance between accuracy and efficiency.

\subsection{Baseline Models}
\label{sec:baseline models}
We compare the proposed approach with several alternatives to demonstrate the advantages and disadvantages of different approaches. The baseline models we use are briefly introduced as follows.

\paragraph{Expert Policy} The policy is designed by domain experts, and it allocates actions to customers according to different user types within the constrained action set. The expert policy is used as a default policy by the platform.

\paragraph{Rainbow}
Another RL model that we compare our model with is the Rainbow model proposed by~\citet{hessel2018rainbow}. It combines multiple extensions of DQN~\citep{bellemare2017distributional,fortunato2017noisy,mnih2016asynchronous,schaul2015prioritized,van2016learning,wang2015dueling} as an integrated RL agent and outperforms the models with each component separately. We replace BCQ with Rainbow and use the Q values learned from Rainbow as input $q_{ij}$ in the linear programming model.

\paragraph{Supervised Learning}
We use three supervised learning methods, 
logistics regression, Gradient Boost Decision Tree (GBDT), and Xgboost~\cite{chen2016xgboost} 
as baseline models. These methods are in the same framework and contain two components: (1)
a reward prediction model to learn user behavior (whether or not the user will log in on the next day given the state and action), and (2) an action selection model to select the best action according to the estimated reward from the reward prediction model.

Specifically, the reward $r_t$ for $(s_t, a_t)$ is calculated by $ r_t(s_t,a_t)=f_r(s_t, a_t)$, where $f_r$ is a supervised learning model. As $r_t$ denotes whether or not the user logs in on the next day, this supervised learning model tries to solve a classification problem. The supervised learning model learns from the data to predict the reward as accurately as possible. 

After the reward prediction model $f_r$ is trained, we use it to generate a policy: 
\begin{equation}
\label{eq:supervised learning policy}
\pi(a_t|s_t) =   \left\{
\begin{array}{ll}
      1 & a_t =  {\argmax}{arg\,max}_{a_t \in \mathbf{A}}f_r(s_t, a_t) \\
      0 & \text{otherwise}
\end{array} \right.
\end{equation}

Using supervised learning to choose actions is actually a rather simplistic method as it does not consider the long-term return or the influence of current action on future state and reward. However, it is broadly practiced in the industry as it is easy to implement and interpret.

We also combine the supervised learning algorithms with LP. Instead of using Q values in the linear programming model, we use the probability of retention predicted by the classification model $f_r$.

\section{Experiments}
\label{sec:experiments}
In this section, we conduct empirical evaluations on real-world data in both online and offline manners to demonstrate that our solution can effectively generate a robust policy under budget constraints compared with baseline models. We first describe our data and experiment settings and then compare the overall performance of the proposed method with other baselines in offline experiments. Next, we show the performance of the proposed method in the online environment with real traffic. The results demonstrate that our solution can effectively generate a robust policy under the budget constraints compared with baseline models. \\

\subsection{Real-World Dataset}
\label{sec:dataset}
We investigate the performance of BCRLSP on a large-scale real-world dataset. We collaborate with Alibaba Inc. and collect data from the Taobao Special Offer Edition App. This real-world data is used to train and evaluate our framework in an offline manner. The dataset comprises 1.4 million records for a continuous one-month period in December 2020.
We split the interactions into the training, validation, and test sets dis-jointly according to different time ranges. The first 2 weeks are used for training, and the remaining 2 weeks are used for offline testing.

\paragraph{State} To accurately depict the pattern of next-day retention, we need a set of informative features (i.e., state variables) for a given cash bonus collection record. Given our research agenda, our features should be able to capture the contextual information and behavioral information that is either stable over time (i.e., static states) or change with time (i.e., dynamic states). Contextual information includes user demographics, such as gender, age, and location, which portray static states that do not vary over time. Behavioral information includes user behavior variables that mainly capture the dynamic states of users.

User behavior variables can be further categorized into two subgroups, frequency and monetary states. 
Frequency states describe how frequent users claim and redeem the cash bonuses and whether they stay in the app given the cash bonuses. Monetary states characterize the monetary values of cash bonuses and order payments. 
All user behavior variables are created at five different historical time windows (1 day, 3 days, 5 days, 7 days, 15 days).

\paragraph{Action} We have 12 values of cash bonuses in total, so our action space has a size of 12. As explained in Section~\ref{sec:context}, these 12 actions are split into two sets: 10 normal cash bonuses, 0.65, 0.67, 0.71, 0.75, 0.79, 0.83, 0.87, 0.94, 1.01, and 1.05 Yuan, and 2 super cash bonuses, 1.72 and 1.82 Yuan. We include all 12 possible actions in our action set for training, but put a constraint on the feasible actions when making predictions. 

\paragraph{Reward} As mentioned in Section~\ref{sec:mdp}, reward is a 0-1 value indicating whether or not the user logs in the next day. The details of the data and feature generation framework can be found in Supplementary Material.

\subsection{Experiment Settings}
\label{sec:experiment settings}

For both BCQ and Rainbow algorithms, we use a 2-layer fully connected neural network with 128-dimension hidden states in the user behavioral model. We use a threshold of $\xi=0.3$ to eliminate actions when selecting the optimal action using BCQ.\footnote{We tried different thresholds of $\xi=0.3,0.4,0.5,0.6$ and the results are similar.} We use the exploration rate $\epsilon=0.01$ to assign a random action in the exploration subsample.\footnote{We tried different exploration rates of $\epsilon$ between 0.01 and 0.1 and $\epsilon=0.01$ shows the best results.} The discount factor 
is set to $\gamma=1$ as the company values the user retention in the whole trajectory and does not discount the rewards across different time periods.\footnote{Each trajectory has a finite length with the maximum length = 4, so $\gamma$ can be set to 1.} 
Training on 6 CPU cores, 30 GB of memory, and 2 GPU Tesla V100 took roughly 5 hours for the RL models.
The hyper-parameters in the baseline models are set as follows: We use a grid search to find the best hyper-parameters in each supervised learning model in the holdout validation dataset. 
For GBDT: learning rate = 0.1, max depth =3, number of estimator = 500;
For Xgboost: learning rate = 0.05, max depth = 5, number of estimator = 500. 

We launch an offline experiment to calculate the evaluation results based on the approach proposed by~\citet{fonteneau2013batch}. Specifically, we use all the states in the test dataset and consider them as users interacting with the agent. Then, given a state in the dataset $s_t$, the agent generates an action $\hat a_t$ according to its policy. Then we find the records in the dataset where the actual action given by the behavior policy matches the estimated action by the agent: $a_t=\hat a_t$. Denote these matching records as $\mathcal{M} =  \{\tau_1,\dots,\tau_m\}\subset\mathcal{D}$, where $\tau_i=\{(s_t^i,a_t^i,r_t^i)\}_{t=1}^{T}$. 



\paragraph{Evaluation Metrics}  The goal of the budget-constrained incentive marketing is to maximize the total retention rate under a fixed budget. Therefore, we evaluate the performance of agents in $\mathcal{M}$ with the following two metrics:
(a) Average retention rate: 
\begin{equation}
\label{eq:average retention rate}
\text{Ret} = \frac{1}{\sum_{j=1}^{m} T_j}\sum_{k=1}^{m} \sum_{t=0}^{T_j} r_{t}^j
\end{equation}
where $T_j$ denotes the length of the trajectory $\tau_j$; 
(b) Average cost:
\begin{equation}
\label{eq:average cost}
\text{AvgCost} = \frac{1}{\sum_{k=1}^{m} T_j}\sum_{j=1}^{m} \sum_{t=0}^{T_j} c_{t}^j
\end{equation}

\subsection{Offline Experiment Results}
\label{sec:offline experiment results}
The results of offline experiments are shown in Table~\ref{table:comparison-table}. Our proposed BCRLSP model and all baseline models are able to maintain the same average cost because the LP model is set under a fixed budget of 0.87 Yuan per customer per day.
Under the same budget setting, BCRLSP model outperforms all baseline models and achieves the highest retention rate. Compared with the second-best model (GBDT), the retention rate of BCRLSP model increases from $36.15\%$ to $37.22\%$, an improvement of $2.96\%$. We train the models using the same data and hyper-parameters for 10 times, and the retention rates change within the range of $0.1\%$.

\begin{table}[t]
\centering
\begin{tabular}{lcc}
\toprule
 Algorithm      & Ret  & AvgCost \\
\midrule
 BCRLSP     & \textbf{37.22\%} & 0.87 \\
 Rainbow    & 29.05\%  &   0.87 \\
 GBDT       &36.15\% & 0.87 \\
 LR         &34.43\% &  0.87 \\
\bottomrule
\end{tabular}
\caption{Offline experiment results}
\label{table:comparison-table}
\end{table}

\subsection{Online Experiment Results}f
\label{sec:online experiment results}
We also evaluate the BCRLSP framework for a week in the online environment, i.e., real-world cash bonuses on the check-in page in the Taobao Special Offer Edition App. 
The comparison baseline models are the expert policy, Rainbow, GBDT, and logistics regression. To assess the effectiveness of the proposed framework, the platform assigns $92\%$ of the online traffic to the expert policy and $2\%$ of the online traffic to each of the BCRLSP, Rainbow, GBDT, and logistic regression models, which is big enough considering the overall amount of user interactions. Table \ref{table:online-table} lists the performance of the two main online metrics. 

Compared with the expert policy, $0.53\%$ growth on retention rate exhibits that our proposed framework increases the retention rate of the platform when put into production.
Note that during the online experiment, the firm decided to increase the budget to 1.60 Yuan, hence we can see an increase in both the average cost and the retention rate in the online experiment results compared with offline ones.


\begin{table}[t]
\centering
\begin{tabular}{lcc}
\toprule
 Algorithm      & Ret  & AvgCost \\
\midrule
 BCRLSP     & \textbf{50.81\%} & 1.60 \\
 Rainbow    & 50.57\%  &   1.60 \\
 GBDT       & 50.69\% & 1.60 \\
 LR         & 50.51\% &  1.60 \\
 Expert policy & 50.54\% & 1.60 \\
\bottomrule
\end{tabular}
\caption{Online experiment results}
\label{table:online-table}
\end{table}

\section{Conclusions}
\label{sec:conclusions}
In this paper, we propose the BCRLSP framework to solve the sequential targeted promotion problem. We combine the offline deep reinforcement learning algorithm with the linear programming model to sequentially target customers with cash bonuses under the budget constraint. As the promotion cost constraint is vital to the company, the RL model is trained offline using logged data to avoid the risk of over-spending. After getting the Q values of actions corresponding to the states of each customer, we maximize the total value across all customers under the constraint of the average cost using the online linear programming model. Compared with other reinforcement learning and supervised learning algorithms, our approach achieves better performance in both customer retention rate and average cost. 

Our work enables companies to better target customers using sequential promotion, which will result in higher customer engagement and loyalty in the long term. Companies will benefit from this research as they can build customer loyalty in a more cost-efficient way. At the same time, this does not mean that customers are put at a disadvantage, as they receive cash bonuses from companies. In other words, our work creates a win-win outcome for companies and customers by reducing the information friction and improving the total surplus in the market.

\bibliographystyle{ACM-Reference-Format}
\bibliography{main}

\end{document}